\def\cameraready{1}
\def\arxivversion{1}
\documentclass{article}

\ifdefined\cameraready
  \usepackage[final,dblblindworkshop]{neurips_2026}
  \workshoptitle{Interpretability as a Science}
\else
  \usepackage{neurips_2026}
\fi
\usepackage{amsmath,amssymb}
\usepackage{booktabs}
\usepackage{colortbl}
\usepackage{float}
\usepackage{graphicx}
\usepackage{tabularx}
\usepackage{wrapfig}
\usepackage[most]{tcolorbox}
\usepackage{xcolor}
\usepackage{hyperref}
\hypersetup{hidelinks}
\ifdefined\cameraready
  \hypersetup{pdftitle={Do Knowledge- and Reasoning-Oriented Benchmarks Induce Stable Representation Signatures in LLMs?},
    pdfauthor={Parsa Rahimi Noshanagh and Azim Dehghani Amirabad}}
\fi
\usepackage{url}
\usepackage{pgfplots}\pgfplotsset{compat=1.18}
\usetikzlibrary{arrows.meta,calc,positioning,backgrounds}
\usepackage{pifont}

\title{Do Knowledge- and Reasoning-Oriented Benchmarks Induce Stable
Representation Signatures in LLMs?}

\ifdefined\cameraready
    \author{
    Parsa Rahimi Noshanagh\\
    EPFL, Idiap Research Institute\\
    \texttt{parsa.rahiminoshanagh@epfl.ch}
    \And
    Azim Dehghani Amirabad\\
    Independent Researcher\\
    \texttt{azim.dehghani@gmail.com}
    }
\fi

\newcommand{\K}{\mathcal{S}_{K}}
\newcommand{\R}{\mathcal{S}_{R}}
\newcommand{\norm}[1]{\left\lVert#1\right\rVert}
\definecolor{fisherblue}{HTML}{3155B7}
\definecolor{probeorange}{HTML}{D87400}
\definecolor{geometryteal}{HTML}{16828A}
\definecolor{rankgray}{HTML}{62666A}
\newcommand{\fishermetric}{\textcolor{fisherblue}{\textbf{Mean shift}}}
\newcommand{\probemetric}{\textcolor{probeorange}{\textbf{Probe}}}
\newcommand{\geometrymetric}{\textcolor{geometryteal}{\textbf{Geometry}}}
\newcommand{\rankmetric}{\textcolor{rankgray}{\textbf{Rank}}}
\tcbset{
  mygoal/.style={
    colback=yellow!10,
    colframe=black!30,
    boxrule=0.6pt,
    arc=1mm,
    left=0.1pt,
    right=0.1pt,
    top=0.1pt,
    bottom=0.1pt,
    enhanced,
    fonttitle=\bfseries,
    coltitle=black,
    attach title to upper,
    after title={\space}
  }
}

\begin{document}
\maketitle

\begin{abstract}
Internal signatures could support monitoring, adaptive computation, and mechanistic
analysis, but benchmark prompts can be separable without any shared computational
distinction. We test whether a prespecified knowledge/reasoning label defines a reusable
internal signature in five contemporary 7B to 31B models. We treat the \emph{\textbf{task} as the unit of
inference}: probes hold out complete tasks, labels are permuted across tasks, and the best layer and stream
are reselected for every permutation. Raw probes recover an eight-versus-eight-subject
MMLU partition in all five models (macro-F1 $0.78$--$0.81$), but prompt text predicts it
nearly as well: under an identical probe configuration, the activation advantage is only
$0.017$--$0.070$ macro-F1 and is not significant. At the prespecified $0.01$ level, lexical
adjustment removes significance while the adjusted states still identify individual
subjects, and the selected site changes across task panels and selection rules.
Additionally, two released base-training trajectories show no common direction. These pooled layerwise measures recover the task label but identify no
transferable knowledge/reasoning signature at the tested sensitivity, which cannot exclude
moderate effects. The
resulting protocol gives future internal monitors and intervention studies a sequence of
tests to pass.
\ifdefined\cameraready
Project page: \url{https://parsa-ra.github.io/lmbehave/}.
\fi
\end{abstract}

\begin{tcolorbox}[mygoal,title=Study at a glance:]
\\
\textbf{Question:} Does a knowledge/reasoning benchmark label identify model processing
beyond task and prompt identity? \\
\textbf{Findings:} Raw states predict the label, but their
matched gain over prompt text is not significant, and the selected sites do not transfer
consistently. \\
\textbf{Use:} The protocol tests representation signals before they guide
monitoring, adaptive computation, or causal intervention.
\end{tcolorbox}

\section{Introduction}

Benchmark suites and downstream analyses often use contrasts between knowledge
retrieval and multi-step reasoning
\citep{hendrycks2021mmlu,wang2024mmlupro,sprague2024musr,song2026demystifying}.
If these contrasts induced stable internal signatures, they could motivate latent-state
monitors \citep{feng2025propositional,mckenzie2025highstakes}, adaptive-compute policies
\citep{liu2023dejavu}, or mechanistic hypotheses
\citep{hendel2023taskvectors,todd2024functionvectors}. Decoding a label is not sufficient,
however: a model can encode task identity, vocabulary, and format without implementing one
shared \emph{knowledge} or \emph{reasoning} computation
\citep{hewitt2019control,belinkov2022probing,kumar2022probing,sahoo2026format}.

These failure modes matter. A monitor driven by topic or format can appear
accurate on one benchmark and fail when prompts or tasks change. A layer that maximizes a
contrast on one task set can be a poor intervention target
\citep{sahoo2026format,hase2023localization}. A candidate signature should therefore
generalize across held-out tasks, add information beyond the input, and remain stable across
task panels, benchmarks, and site-selection rules (selectors) before it motivates monitoring or causal intervention.

We ask four questions about whether the benchmark label, at this coarse level, can serve as a reusable
internal signal. \textbf{First}, can intermediate states predict the label of a task that was
entirely excluded from probe fitting? \textbf{Second}, do they predict it better than the prompt text
under a matched estimator? \textbf{Third}, does the selected site transfer across panels and
selectors? \textbf{Fourth}, as a separate training diagnostic, does the measured profile change
consistently as accuracy on downstream benchmarks improves? The first three separate recovery of the task label from
the stronger evidence that a monitor, router, or intervention target needs. The fourth asks
whether the measured profile changes consistently during training, which determines whether
it could serve as a training-time signal for future models whose stage schedules differ
from those common today.

\begin{figure}[t]
\centering
\resizebox{0.985\linewidth}{!}{\input{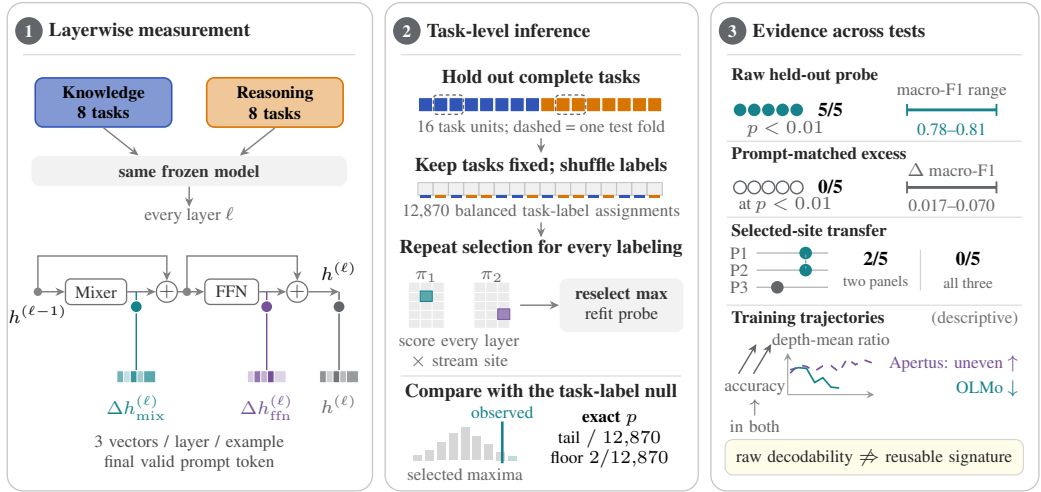}}
\caption{Study design and results. Panel 1: token-mixing and feed-forward updates are read
before residual addition, and the residual state after it, all at the final valid prompt
token (shading is schematic). Panel 2: held-out-task inference exhaustively permutes task
labels ($\pi_1,\pi_2$: two permuted labelings) and reselects the layer--stream site. Panel
3: raw probes recover the partition in every model, but the matched gain over prompt text is
not significant, sites do not transfer consistently across the three task panels (P1 MMLU-Pro, P2 MMLU, P3 heterogeneous),
and the two training trajectories disagree.}
\label{fig:overview}
\end{figure}

Our central finding (Figure~\ref{fig:overview}) is a gap between \emph{decodability} and
\emph{specificity}. Raw probes recover the label in all five models, but their matched advantage over prompt features is
not significant, the selected site depends on the panel and selector, and the two training
trajectories do not share a direction. On these task sets, the tested layerwise summaries identify no reusable knowledge/reasoning
axis, although moderate effects remain possible. Our contributions are:
\begin{itemize}
    \item We formulate task-set separation as an exact grouped randomization test. Probes
    are tested on held-out tasks, labels are permuted across whole tasks, and site selection
    and probe fitting are repeated for every permutation
    (Section~\ref{sec:measurements}).
    \item Across five 7B to 31B checkpoints, raw selected-site probes are significant in every
    model ($p=0.00155$--$0.00420$), whereas the matched activation-minus-prompt differences
    are not ($p=0.177$--$0.440$). On an arbitrary control partition, a naive probe that shares
    tasks between training and test data is significant where the task-held-out probe is not
    (Section~\ref{sec:cross-model-results}).
    \item We test whether the selected site is stable across three task panels and against three covariance-whitened selectors, and we study two released base-training
    trajectories.
    Neither the selected sites nor the trajectory directions are stable
    (Sections~\ref{sec:longitudinal} and~\ref{sec:cross-model-results}).
\end{itemize}
Our contribution is an evaluation protocol: task-level holdout and permutation, a matched
prompt comparison, transfer tests, and a detection calibration, which together separate
task-label decodability from the evidence that reusable internal monitoring requires.

\subsection{Related work and scope}

Behavioral evaluation, observational decoding, and intervention answer distinct questions:
they measure outputs, recoverable information, and output sensitivity under manipulation,
respectively \citep{hewitt2019control,belinkov2022probing,meng2022locating,conmy2023acdc}.
Latent-state methods can recover propositions or deployment-relevant labels
\citep{burns2023discovering,feng2025propositional,mckenzie2025highstakes}, while probes can
also learn dataset regularities \citep{pimentel2020information}. \citet{sahoo2026format} show
that format controls weaken a reasoning-mode probe, and they add a causal steering
experiment. We test a different binary partition without reproducing their
intervention, and we add task-level exchangeability, exhaustive permutation,
selection-aware depth inference, five contemporary checkpoints, and two training trajectories.
Concurrent contamination-probing work likewise emphasizes nuisance-matched baselines and
detection calibration \citep{braun2026excess}.

Task and function vectors make raw task separation plausible
\citep{hendel2023taskvectors,todd2024functionvectors}. Our narrower question is whether
these task differences align with one transferable knowledge/reasoning direction.
Intervention studies report shallower dependence for knowledge retrieval and deeper
dependence for reasoning \citep{song2026demystifying}, or selective recall-versus-reasoning
effects on controlled puzzles \citep{fartale2025disentangling}. These task-specific causal
findings motivate our test, but they neither imply nor require a benchmark-independent
passive signature.
Causal tracing restores a clean hidden state in a corrupted run and measures output
recovery; activation patching substitutes a component state between paired runs and measures
the output change \citep{meng2022locating,conmy2023acdc}. Our passive measurements instead
generate hypotheses for such interventions.

We report prompt residualization only as a sensitivity analysis, because successful linear
removal does not prove concept removal \citep{kumar2022probing}. Effective rank diagnoses
variance contraction \citep{roy2007effectiverank}, paired CKA compares geometries
\citep{kornblith2019similarity}, and sparse autoencoders address a different, feature-level
question \citep{huben2024sae}. Finally, the benchmark categories are operational proxies:
MMLU and MMLU-Pro subjects mix factual recall and inference
\citep{hendrycks2021mmlu,wang2024mmlupro}, so our conclusions concern the named task sets and
this measurement protocol.

\section{Problem Formulation and Method}
\label{sec:measurements}

\subsection{Task-level setup and representation sites}

Let
\begin{equation}
  \mathcal T=\K\mathbin{\dot\cup}\R,\qquad
  y_t=\mathbf{1}[t\in\K],\qquad
  \mathcal D_t=\{x_{ti}\}_{i=1}^{n_t},
  \label{eq:task-setup}
\end{equation}
where $\K$ and $\R$ are the sets of knowledge-oriented and reasoning-oriented tasks, $t$
indexes benchmark tasks, and $i$ indexes examples within a task. The label is
therefore defined per task, and examples from the same task are not independent instances of
the knowledge/reasoning contrast. For model $m$, the supported observation sites are
\[
  \mathcal J_m=\{(\ell,s):0\leq\ell<L_m,\;
  s\in\{\mathrm{Mix},\mathrm{FFN},\mathrm{Residual}\}\}.
\]
At layer $\ell$, Mix is the verified token-mixing contribution immediately before residual
addition, FFN is the analogous feed-forward contribution, and Residual is the complete
decoder-layer output after the residual updates. Mix can be softmax attention or another
learned sequence mixer. We treat the three as separately observed streams \emph{without} assuming
that they add up, and we record a branch as missing when its semantics cannot be verified.

Let $H_{ti,j}\in\mathbb R^{N_{ti}\times d_m}$ be the hidden sequence at site
$j=(\ell,s)$ and $a_{ti}$ its prompt mask. Each prompt is represented by its state at the final valid (non-padding) token:
\begin{equation}
  r_{ti}=\max\{u:a_{ti,u}=1\},\qquad
  z_{ti,j}=H_{ti,j}[r_{ti}]\in\mathbb R^{d_m}.
  \label{eq:pooling}
\end{equation}
The primary study measures the prompt (prefill) state at this one position, and a separate
descriptive analysis averages valid generated-token states within relative response-position
bins. The forward hooks only read activations and leave weights and
inference settings unchanged. For architecture mappings and invariance checks, see
Appendix~\ref{app:implementation}.

\subsection{Layerwise estimands}

For each site $j$, a label-blind PCA map $P_j$ retains the smallest basis explaining at
least 95\% of total variance. Let $\tilde z_{ti,j}=P_j(z_{ti,j}-\bar z_j)$, with means
$\mu_{c,j}$ and sample covariance matrices $\Sigma_{c,j}$ for the two task categories
$c\in\{0,1\}$ given by the label $y_t$ (knowledge-oriented tasks have $c=1$). The prespecified
localization statistic is the PCA-space mean-shift ratio
\begin{equation}
  R_j(y)=
  \frac{\norm{\mu_{1,j}-\mu_{0,j}}_2^2}
       {\operatorname{tr}(\Sigma_{1,j})+
        \operatorname{tr}(\Sigma_{0,j})+10^{-12}},
  \qquad
  j^\star(y)=\arg\max_{j\in\mathcal J_m}R_j(y).
  \label{eq:mean-shift}
\end{equation}
Unlike the classical Fisher criterion, this ratio does not whiten features by the inverse
covariance, so it need not rank sites by linear separability. It is recomputed exactly from
per-task sums for every task-label permutation without inverting a covariance; a regularized
covariance-whitened variant is reported as a sensitivity analysis
(Appendix~\ref{app:analysis-details}). We report its numerator, denominator, PCA dimension,
and within-category effective rank so that variance contraction remains visible.

Let the folds $F_1,\ldots,F_K$ partition the complete tasks. For fold $k$, preprocessing and
the classifier are fitted only on examples from tasks outside $F_k$, and pooling the
out-of-fold predictions defines
\begin{equation}
  T_{\mathrm{rep}}(y;j)=\operatorname{MacroF1}
  \left(\{y_t,\widehat g_j^{(-F_{k(t)})}(z_{ti,j})\}_{t,i}\right).
  \label{eq:grouped-probe}
\end{equation}
Probe fitting holds out tasks, but the site $j^\star(y)$ is chosen from all tasks, so the
reported macro-F1 is held out for the probe but not for site choice. The exact test accounts
for this choice by repeating it for every permutation.
The matched prompt comparison reduces the activations and the prompt's character-n-gram
TF--IDF features to 64 components each, fitted within each fold, then standardizes both and
fits the same logistic classifier with the same folds and regularization. Its paired statistic is
\begin{equation}
  \Delta(y)=T_{\mathrm{rep}}^{64}(y;j^\star(y))-
            T_{\mathrm{text}}^{64}(y).
  \label{eq:matched-excess}
\end{equation}
$T_{\mathrm{rep}}$ thus tests recoverability, whether activations predict the label, and
$\Delta$ tests whether they predict it better than the matched prompt text. As a separate
sensitivity analysis, lexical adjustment removes, within each fold, the activation variance
linearly predictable from 64 TF--IDF components of the prompt (lexical-64).

\subsection{Exact task-label inference and study questions}

For $q=|\K|$, the conditional randomization space is
\begin{equation}
  \Pi_q=\left\{y'\in\{0,1\}^{|\mathcal T|}:\sum_t y'_t=q\right\}.
\end{equation}
For statistic $S$, exhaustive task-label inference uses
\begin{equation}
  p_S=\frac{1}{|\Pi_q|}\sum_{y'\in\Pi_q}
      \mathbf{1}\!\left[S(y')\geq S(y)\right].
  \label{eq:exact-null}
\end{equation}
For every $y'$, grouped folds are rebuilt, $R_j(y')$ is recomputed, $j^\star(y')$ is
reselected, and the probe is refit. With eight tasks per category,
$|\Pi_8|=\binom{16}{8}=12{,}870$. Because the statistics are symmetric in the two class
names, the smallest attainable p-value is $2/12{,}870$, well below the prespecified
threshold of $p<0.01$, which was fixed in the frozen analysis manifest before the primary
panel's results were summarized; in the five-versus-five panels only the floor
$1/126=0.00794$ lies below it. Benjamini--Hochberg correction is applied across all
layer--stream sites of a model, and the maximum test accounts directly for site selection
\citep{westfall1993resampling}. Secondary metrics (RBF probes, cosine silhouette,
covariance alignment, paired CKA, and effective rank) are defined in
Appendices~\ref{app:metric-definitions}--\ref{app:analysis-details}, and Appendix
Figure~\ref{fig:metric-illustration} illustrates what each statistic responds to.

This formulation separates four questions. \emph{Recoverability} asks whether
$\max_jR_j(y)$ or $T_{\mathrm{rep}}(y;j^\star(y))$ is significant under
Equation~\ref{eq:exact-null}. \emph{Prompt excess} asks the same of $\Delta(y)$.
For panels $p,p'$, we describe localization transfer by
\begin{equation}
  A_m^{p,p'}=\mathbf{1}[j^\star_{m,p}=j^\star_{m,p'}],\qquad
  \rho_{m,s}^{p,p'}=\operatorname{corr}_{\ell}
  (R_{m,p,\ell,s},R_{m,p',\ell,s}).
  \label{eq:transfer}
\end{equation}
Finally, for training checkpoint $c$, the depth mean
\begin{equation}
  D_{m,s}(c)=\frac{1}{L_m}\sum_{\ell=0}^{L_m-1}R_{m,c,\ell,s}
  \label{eq:depth-mean}
\end{equation}
tracks whether the measured profile changes consistently with accuracy along a released
trajectory.

All four test specified task partitions under a frozen design. Semantic purity,
causal specialization, and pruning safety lie outside their scope.

\section{Experiments}

We begin with the longitudinal analysis, which holds model identity fixed, and then we test
specificity and transfer across five contemporary instruction-tuned models.

\subsection{Longitudinal training studies}
\label{sec:longitudinal}

A stable task-set distinction need not strengthen with accuracy, so association with
training is informative but not necessary. We ask the narrower empirical question of whether
the measured profile changes systematically as performance improves. OLMo 3 and Apertus 1.0
release checkpoints from documented, staged base-training runs \citep{olmo3,apertus2025},
which lets us study this within one model at a time. We evaluate seven OLMo 3 7B and eleven
Apertus 1.0 8B checkpoints, holding examples, prompts, tokenization, pooling, precision,
code, and analysis fixed. Within each trajectory, every checkpoint uses the same three
non-overlapping blocks of 32 examples from each of the same ten MMLU subjects (anatomy,
clinical knowledge, world history, professional law, and astronomy versus abstract algebra,
formal logic, high-school mathematics, college mathematics, and logical fallacies).

\paragraph{Uncertainty.}
Means and sample standard deviations are taken across the three paired blocks; they measure
sensitivity to example selection within a single training run. Greedy prefill
measurement is deterministic, so rerunning generation seeds on the same examples would
reproduce the same states. The probe uses all 251 non-observed balanced task-label
permutations.

\begin{figure}[t]
\centering
\includegraphics[width=\textwidth]{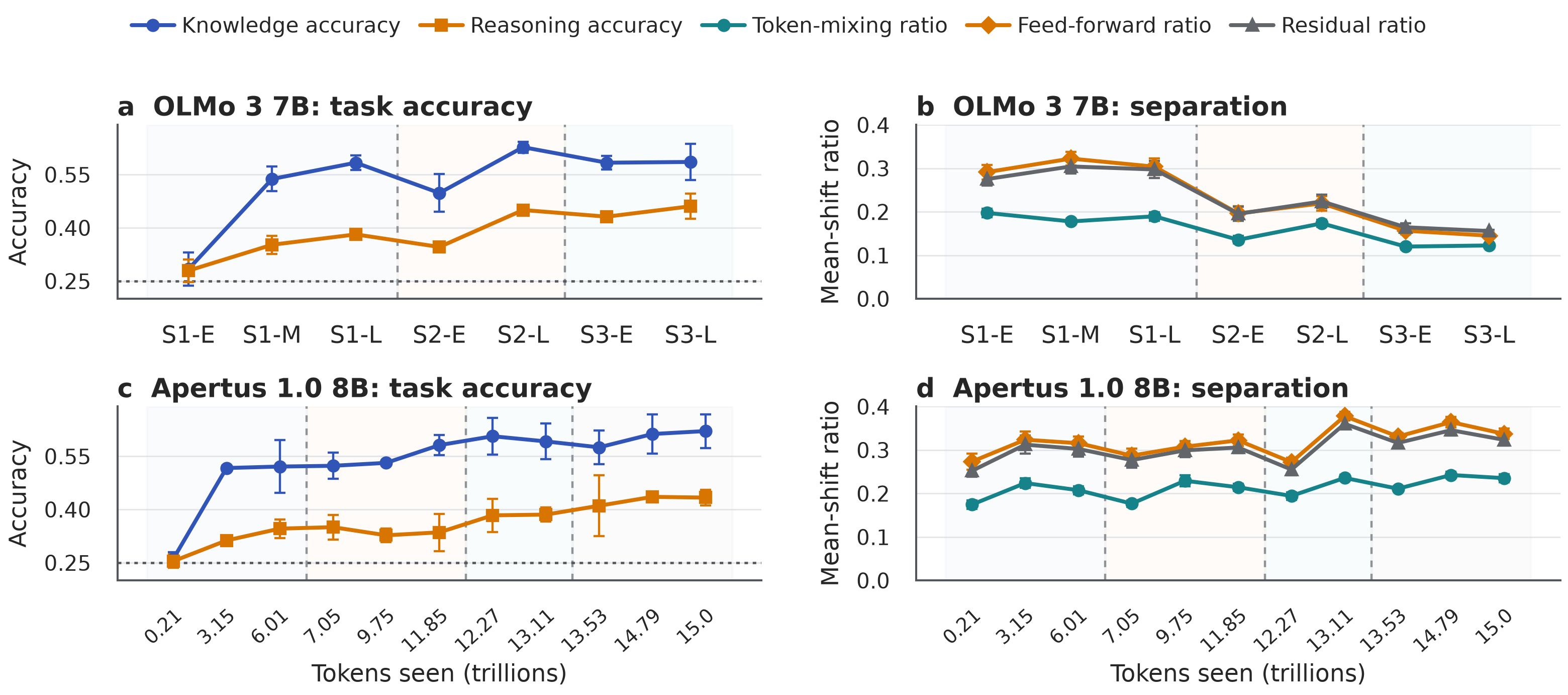}
\caption{Two released base-training trajectories give different endpoint directions.
Accuracy rises in both models. OLMo's depth-averaged mean-shift ratios end below their earliest
values, whereas Apertus's end above them. Neither path is monotonic. Error bars are sample
standard deviations across three paired example blocks (example-sampling variation only). Dashed lines separate documented training stages. OLMo labels use E/M/L for the
early, middle, and late sampled checkpoint within each stage; Apertus labels give trillions
of tokens seen.}
\label{fig:longitudinal}
\end{figure}

For checkpoint $c$ and stream $s$, we use the depth summary in
Equation~\ref{eq:depth-mean}. Because OLMo step counters reset between stages, we report
per-step rates of change only within a stage. Apertus checkpoints are indexed by cumulative
training tokens. OLMo's three documented stages are pretraining, midtraining, and
long-context extension, whereas the Apertus checkpoints span pretraining stages only, so
the two curves cover different stage types. Figure~\ref{fig:longitudinal} shows every sampled checkpoint, and
Table~\ref{tab:longitudinal-results} compares the endpoints.

\begin{table}[t]
\centering
\footnotesize
\caption{Means across three paired example blocks at the first and final checkpoints.
Arrows point from the first to the final checkpoint. Figure~\ref{fig:longitudinal} shows every checkpoint
with sample standard deviations.}
\label{tab:longitudinal-results}
\resizebox{\textwidth}{!}{%
\begin{tabular}{lccccc}
\toprule
Model (checkpoints) &
$\mathrm{Acc}_{K}\,\uparrow$ & $\mathrm{Acc}_{R}\,\uparrow$ &
$D_{\mathrm{Mix}}$ & $D_{\mathrm{FFN}}$ &
$D_{\mathrm{Residual}}$ \\
\midrule
OLMo 3 7B (7) & $0.283\!\rightarrow\!0.585$ & $0.279\!\rightarrow\!0.460$ &
$0.197\!\rightarrow\!0.123$ & $0.291\!\rightarrow\!0.145$ & $0.275\!\rightarrow\!0.156$ \\
\midrule
Apertus 1.0 8B (11) & $0.260\!\rightarrow\!0.621$ & $0.254\!\rightarrow\!0.433$ &
$0.174\!\rightarrow\!0.235$ & $0.273\!\rightarrow\!0.337$ & $0.252\!\rightarrow\!0.323$ \\
\midrule
Reference & $0.250$ chance & $0.250$ chance & \multicolumn{3}{c}{$0$: equal projected means} \\
\bottomrule
\end{tabular}
}
\end{table}

These evaluations retained parsed outputs only. Every OLMo response parses as A--D, and the
Apertus~1.0 parse rate is $0.997$ at its first two checkpoints and $1.000$ thereafter, so the
parsing failures reported for Apertus~1.5 (Section~\ref{sec:sensitivity}) do not drive the
accuracy trends below.

\textbf{OLMo.} From the earliest Stage~1 to the final Stage~3 checkpoint,
knowledge/reasoning accuracy rises from $0.283/0.279$ to $0.585/0.460$ while $D_s$ falls from
$0.197/0.291/0.275$ to $0.123/0.145/0.156$ (token mixing, feed-forward, residual). The mean-shift ratio rises within
Stage~2 but moves in mixed directions within Stages~1 and~3. The other measurements disagree
with one another: linear-probe macro-F1 rises modestly ($0.811/0.806/0.814$ to
$0.837/0.842/0.849$), whereas RBF-probe macro-F1 ($0.798/0.827/0.844$ to $0.792/0.813/0.807$) and
silhouette ($0.140/0.194/0.188$ to $0.096/0.108/0.114$) give no sign of increasingly distinct
geometries. Within-group effective-rank fractions rise ($0.669/0.654/0.700$ to
$0.694/0.683/0.716$), so the depth mean falls without a decrease in the
effective-rank fraction.

\textbf{Apertus.} From 210B to 15T tokens, accuracy rises from $0.260/0.254$ to
$0.621/0.433$, and $D_s$ rises from $0.174/0.273/0.252$ to $0.235/0.337/0.323$, though not
monotonically. At the Stage~3--4 boundary both accuracies rise ($0.581/0.335$ to
$0.606/0.383$) while all three mean-shift summaries fall ($0.214/0.322/0.306$ to
$0.194/0.272/0.255$). Linear-probe macro-F1 barely moves ($0.798/0.783/0.790$ to $0.806/0.792/0.795$),
RBF-probe macro-F1 has mixed endpoint directions, and effective-rank fractions stay near
$0.64$--$0.67$, so the depth mean rises without a decrease in this rank summary.

Because the checkpoints are serially dependent and span different training stages, we treat
these results as trajectory descriptions rather than independent-sample tests. The two
trajectories share no common direction: OLMo depth means fall, Apertus means rise, and both
change non-monotonically between checkpoints. Each is one base-training run, and $D_s$
differs from the selected-site statistic, so these observations cannot support
population-level conclusions about training trends.

We next test whether contemporary instruction-tuned checkpoints nevertheless show recoverability,
incremental prediction beyond prompt text, and transferable localization.

\subsection{Cross-sectional design}

\paragraph{Models and task panels.}

We study OLMo 3 7B Instruct \citep{olmo3}, Apertus 1.5 8B
\citep{apertus2025,apertus15modelcard}, Qwen 3.5 9B and 27B \citep{qwen35}, and Gemma 4
31B IT \citep{gemma4}. Apertus 1.5, Qwen 3.5, and Gemma 4 are multimodal models evaluated here on text only.
Primary comparisons are within model: raw representations and absolute separation values
are never pooled across architectures or hidden widths. Full configurations appear in
Appendix~\ref{app:experimental-details}.

The heterogeneous cross-benchmark panel places five knowledge-oriented tasks against five
reasoning-oriented tasks. All inputs are text, but task family, domain, prompt length,
number of choices, and answer type remain confounded with the label, so we use this panel
only as a transfer test of the task partition.

We use two distinct single-prompt panels. The first is an MMLU-Pro pilot with five subjects
per group and a common ten-option prompt: biology, health, history, law, and economics are
contrasted with mathematics, physics, computer science, engineering, and philosophy
\citep{wang2024mmlupro}. The primary MMLU panel is drawn from a different dataset. It has
eight subjects per group in MMLU's four-option format and contrasts anatomy, astronomy,
clinical knowledge, world history, professional law, high-school biology, medical genetics,
and prehistory with abstract algebra, college mathematics, formal logic, high-school
mathematics, logical fallacies, college physics, conceptual physics, and high-school
statistics \citep{hendrycks2021mmlu}. Using 16 subjects increases the number of distinct
class-symmetric task-label permutations from 126 to 6,435.

Within each panel, all tasks share one prompt, parser, and generation procedure, but the two
panels differ in dataset, subject composition, prompt text, and answer space. The category
labels are comparative proxies: clinical and physics questions can require both acquired
facts and inference. The primary condition uses 100 examples per task, the largest common
count among the selected MMLU subjects, and requests one answer label with a four-token
ceiling. We use these protocol-specific scores only to check that the evaluation elicits valid task
responses.

We test cross-panel transfer because a reusable monitor or intervention target should work
beyond one task panel. A failure to replicate limits that use.

\paragraph{Experimental controls.}

Across models we fix dataset and model revisions, document IDs, raw task prompts,
preprocessing, batching policy, numerical precision, backend, generation settings, and
seeds; the batch size is fixed within each model (Table~\ref{tab:configurations}). Each
instruction model uses the chat template shipped with its tokenizer, so the fully rendered
input is model-specific, and Qwen 3.5 and Apertus 1.5 run with their templates' thinking
mode disabled. Primary runs use deterministic decoding. Unavailable
output lengths are recorded as missing values, and the primary prefill analysis does not
condition on downstream correctness or response length.

\paragraph{Analysis plan and inference rules.}

Within each model, the prespecified localization site is the layer--stream site with maximum
PCA-space mean-shift ratio $R_{\ell,s}$. A second maximum excludes residual states and selects
between the token-mixing and feed-forward updates. We treat either site as exploratory
unless it replicates when site selection is also repeated inside each fold, or in an
independent model.

Mean-shift maxima, counts of significant layer--stream sites, and raw probes test
recoverability. The matched representation-minus-prompt statistic, separate from the
prespecified localization statistic, tests prediction beyond prompt text and is the relevant
evidence for that narrower claim, and the heterogeneous panel tests transfer. RBF probes, silhouette, covariance alignment, and post hoc pooling or subgroup
analyses are sensitivity analyses or descriptive outcomes, and separately selected maxima
are not combined.

\subsection{Cross-model results}
\label{sec:cross-model-results}

Each model condition has 100 examples from each of the 16 MMLU subjects, or 800 per group
at every layer and stream.

\textbf{Mean-shift maximum.} We recompute every layer--stream mean-shift ratio and reselect
the maximum under all 12,870 task-label permutations. The exact test rejects the task-label
null at $p<0.01$ for Apertus~1.5 and both Qwen sizes, but not for OLMo or Gemma
(Table~\ref{tab:full-results}). An alphabetical eight-versus-eight split, used as a negative control, is not
significant in any model ($p=0.147$--$0.269$). Many individual layer--stream
sites are significant, but these sites are dependent, so their counts do not measure
model-level effect size (Appendix~\ref{app:implementation}). Every primary maximum is a
token-mixing or feed-forward site, so the update-only and overall mean-shift maxima
coincide in this panel.

\textbf{Selected-site probe.} When the mean-shift site is reselected and the task-held-out
linear probe is refit for every task-label permutation, macro-F1 is $0.78$--$0.81$, and the
exact test rejects the task-label null in all five models ($p\leq0.0042$;
Table~\ref{tab:full-results}). This includes OLMo and Gemma, whose mean-shift maxima are not
significant at $p<0.01$.

\textbf{Prompt text alone.} TF--IDF probes on the recorded input prompts, with vocabulary,
weights, and classifier refit inside every held-out-task fold and balanced task-label permutation, reach
macro-F1 $0.701$ for word 1--2-grams ($p=0.00870$) and $0.768$ for character-within-word
3--5-grams ($p=0.000777$). The task partition is therefore predictable from the prompt text
alone under the same task-level inference.

\textbf{Matched excess over prompt text.} With the matched estimator of
Equation~\ref{eq:matched-excess}, which reselects the activation site under every
permutation, the prompt probe obtains
macro-F1 $0.755$ ($p=0.00140$). Representation macro-F1 is $0.773$--$0.825$, but the paired
advantage is only $0.017$--$0.070$ and is not significant for any model under the 12,870
balanced task-label permutations ($p=0.177$--$0.440$; Table~\ref{tab:full-results}). For an
alphabetical negative-control split with eight subjects per group, representation and excess
p-values are $0.228$--$0.313$ and $0.790$--$0.861$.

Across all task-label permutations, the null distribution of $\Delta$ has mean
$0.006$--$0.017$, SD $0.057$--$0.067$, and 95th percentile $0.108$--$0.129$, and every
observed $\Delta$ lies inside this distribution (Appendix
Table~\ref{tab:review-validation}). Replacing activation PCA-64 with
fold-local 95\%-variance PCA gives deltas of $0.025$--$0.050$ that are likewise not
significant ($p=0.257$--$0.466$), which argues against the fixed activation width of 64 as
the explanation. In an across-model sensitivity analysis with Holm correction over the five
models in each analysis condition, all raw probes remain significant at $p<0.01$, whereas no
model in the matched-excess or lexical-64 condition falls below $0.05$ after correction.

\textbf{Unit of inference.} On the alphabetical negative-control split, an OLMo probe whose training and
test sets both contain examples from every task obtains macro-F1 $0.802$ ($p=0.0001$ under
9,999 naive sample-label permutations), whereas task holdout gives $0.574$ ($p=0.228$): treating examples rather than tasks as the
unit, in both splitting and permutation, inflates both the score and its significance
(Appendix~\ref{app:analysis-details}).

\begin{table}[!ht]
\centering
\footnotesize
\setlength{\tabcolsep}{3.2pt}
\caption{Primary MMLU inference and cross-panel localization. Shading marks mean-shift
maxima with $p<0.01$. Matched $\Delta$ uses identical 64-feature estimators and reselects
the site for every permutation; lexical-64 refits the lexical adjustment within each fold. Sites are
zero-indexed layer / stream. The
smallest attainable p-value is $2/12{,}870\approx1.55\times10^{-4}$ for MMLU and $0.00794$ for
MMLU-Pro. Heterogeneous-panel sites are
descriptive, and all sites come from the mean-shift selector.}
\label{tab:full-results}
\resizebox{\textwidth}{!}{%
\begin{tabular}{lccccc}
\toprule
Model & MMLU site / \textcolor{fisherblue}{mean-shift ratio $\uparrow$} &
\textcolor{fisherblue}{max $p\downarrow$} &
\textcolor{probeorange}{raw macro-F1$\uparrow$ / $p\downarrow$} &
\textcolor{probeorange}{matched $\Delta$ macro-F1 / $p\downarrow$} &
\textcolor{probeorange}{lexical-64 macro-F1$\uparrow$ / $p\downarrow$} \\
\midrule
OLMo 3 7B & 3 / FFN / 0.221 & 0.0721 & $0.781/0.00155$ & $+0.0476/0.2897$ & $0.666/0.01197$ \\
\rowcolor{yellow!10} Apertus 1.5 8B & 0 / FFN / 0.284 & 0.00435 & $0.804/0.00186$ & $+0.0695/0.1770$ & $0.613/0.06605$ \\
\rowcolor{yellow!10} Qwen 3.5 9B & 2 / FFN / 0.616 & 0.00109 & $0.805/0.00186$ & $+0.0468/0.2909$ & $0.689/0.01476$ \\
\rowcolor{yellow!10} Qwen 3.5 27B & 1 / token mixing / 0.667 & 0.000777 & $0.796/0.00420$ & $+0.0172/0.4396$ & $0.655/0.04165$ \\
Gemma 4 31B & 23 / FFN / 0.499 & 0.0145 & $0.805/0.00404$ & $+0.0511/0.2968$ & $0.698/0.02020$ \\
\bottomrule
\end{tabular}
}

\vspace{3pt}
\begin{tabular}{lccc}
\toprule
Model & MMLU-Pro site / $p$ & MMLU site & Mixed-task site \\
\midrule
OLMo 3 7B & 3 / FFN / 0.0159 & 3 / FFN & 5 / token mixing \\
Apertus 1.5 8B & 0 / FFN / 0.0238 & 0 / FFN & 8 / token mixing \\
Qwen 3.5 9B & 1 / token mixing / 0.0317 & 2 / FFN & 11 / token mixing \\
Qwen 3.5 27B & 11 / FFN / 0.0317 & 1 / token mixing & 5 / FFN \\
Gemma 4 31B & 12 / token mixing / 0.0238 & 23 / FFN & 14 / FFN \\
\bottomrule
\end{tabular}
\end{table}

\textbf{Localization.} Only OLMo and Apertus select the same mean-shift site in both
single-prompt panels, and no model selects the same site in all three (Table~\ref{tab:full-results}).
Between MMLU-Pro and the heterogeneous panel, per-stream Pearson correlations of the depth
profiles range from $-0.047$ to $0.922$, yet no model retains its maximizing site, so a
similar depth profile can coexist with a different maximum.
A ridge-regularized, covariance-whitened variant of the mean-shift ratio (three ridge
strengths; Appendix~\ref{app:analysis-details}) selects a site other than the mean-shift site in all 15
model--ridge conditions, although every whitened maximum is significant after Holm
correction ($p_{\mathrm{Holm}}\leq0.0094$). Localization therefore depends on the selector.
The heterogeneous-panel sites are descriptive because site selection was not repeated
within their probe tests.

\subsection{Sensitivity analyses and causal scope}
\label{sec:sensitivity}

\textbf{Prompt-feature adjustment.} Regressing six scalar prompt features, such as token
count, out of the activations removes every BH-significant linear-probe site in MMLU-Pro (81, 70, 0,
92, and 88 sites per model before adjustment, none after) but leaves significance in four
of the five primary MMLU models. Lexical-64 adjustment lowers primary macro-F1 by $10.7$--$19.1$
points and gives $p=0.012$--$0.066$: none is below $0.01$, four are below $0.05$, and 32
components give the same result. The adjustment is partial and cannot establish mediation:
the adjusted states retain 31--54\% of
between-task variance, 57--76\% of total variance, and strong 16-way task prediction under
stratified example folds (macro-F1 $0.765$--$0.961$, chance $0.0625$;
Appendix~\ref{app:analysis-details}).

\textbf{Detection calibration.} To estimate power, we inject a known shift between two
arbitrary task groups along ten random directions at one fixed OLMo site and rerun the
lexically adjusted probe. It detects 0/10 directions at standardized mean differences of
$0.5$--$2$ and 5/10 at $4$, so moderate signals could go
undetected (Appendix Table~\ref{tab:probe-calibration}). A separate selection-aware control
on the alphabetical partition appends a synthetic site and repeats the complete
matched-$\Delta$ procedure. The synthetic site is selected for none of the ten directions,
even at a standardized difference of $4$, and none is detected. The fixed-site control
therefore cannot calibrate the full selection-plus-probe procedure. The confounded
multimodal analysis is also not significant at $p<0.01$ (Appendix~\ref{app:multimodal-side}).

\textbf{Correctness.} We omit correctness conditioning because sparse cells and parse validity
of $0.573$ (Apertus) and $0.761$ (Gemma) would confound correctness with format compliance.

\textbf{Response position.} In an OLMo pilot where the model explains after answering (four
tasks per group), the largest prompt-to-response increase over all layer--stream--bin
candidates (layer 31, token mixing, bin 5 of 8) raises the ratio from $0.193$ to $0.479$
($p=0.0286$, the smallest value attainable in this design and above the $0.01$ threshold),
while the depth-averaged ratio falls; the pilot is therefore descriptive (Appendix
Figure~\ref{fig:relative-response}).

\section{Discussion and Conclusion}

\subsection{Implications and applications}

Our experiments separate a signal that is \emph{decodable} from one that is specific and
stable enough to be \emph{reused}. A raw probe can recover the benchmark label while a
matched prompt probe performs nearly as well, and a significant maximum can coexist with
unstable localization. Reporting only probe accuracy or the largest layerwise contrast would
therefore overstate what the measurement supports.

The immediate use is methodological. For monitoring or adaptive computation, the protocol
tests whether an internal readout adds information beyond the input and transfers to new
task compositions. For benchmark design, it tests whether a semantic taxonomy reflects more
than surface form. For mechanistic analysis, it separates sites worth testing causally from
maxima that depend on the panel or selector. Training profiles need replication across
checkpoints and runs before they can serve as development signals. This work specifies and tests the evidence
that a router, monitor, or pruning policy would require; it validates no such system.

\subsection{Limitations}

The taxonomy entangles demand, domain, format, and difficulty. Task holdout leaves these
confounds in place, and 16 tasks are few independent units for a construct-level claim. The single-run trajectories and the
small response pilot cannot estimate population variation. Localization remains conditional
on the selector and pooling, and mean pooling, capacity-aware probes, and reliable nonlinear
probes remain to be tested. The selection-aware synthetic control does not establish useful power even
for the largest tested shift, so a non-rejection cannot be read as equivalence or as the
absence of a moderate signal. Finally, the design was not publicly preregistered, and hook
invariance does not establish deployment overhead or robustness.

\subsection{Conclusion}

Raw selected-site probes recover the task partition in all five models, and mean-shift
maxima are significant in three. However, the matched prompt advantage is not significant,
no model reaches the prespecified $p<0.01$ level after lexical adjustment, the selected sites
vary, and the training trajectories disagree. Moderate effects and processing differences
remain possible, but on these task sets pooled layerwise measures alone are insufficient
evidence for a reusable computational signature.
Before downstream use, a stronger measure should add prediction beyond matched inputs,
transfer across panels, show calibrated sensitivity, and support causal tests.

\bibliographystyle{plainnat}
\bibliography{references}

\appendix

\section{Implementation and Verification Details}
\label{app:implementation}

The reference implementation is an opt-in component named
\texttt{RepresentationDynamicsTracker} in our extension of \texttt{lmms-eval}.
PyTorch forward hooks are attached to one identified decoder stack and return no replacement
value. Three hooks per zero-indexed layer capture the token-mixing update, feed-forward
update, and residual state. The implementation raises an error when it cannot identify one
decoder stack unambiguously, records whether each update is a verified residual contribution,
uses the attention mask for padding-aware pooling, and removes all hooks during cleanup.

For OLMo 3 and dense Gemma 4, the verified residual contributions are taken after
post-attention and post-feed-forward normalization. Apertus~1.5 and Qwen 3.5 expose verified
token-mixing and feed-forward updates directly. The term ``token-mixing update'' also covers
architectures whose sequence mixer is not softmax attention. Gemma 4 applies a learned scalar
to the complete layer output after the residual additions, so its adjacent residual change
differs from the sum of the two recorded updates. The maximum relative adjacent-state
reconstruction error is $0.011$--$0.042$ for OLMo, Apertus, and both Qwen sizes, versus
$14.7$ for Gemma. This is consistent with the documented Gemma scalar, and no analysis
assumes that the three streams are additive.

\paragraph{Response-trajectory validation.}
The hooks can retain one vector per example, layer, stream, and generated-token update.
Absolute positions are distinguished from prefill and from one another. For
relative-position analysis, every example's uncapped valid updates are split
into eight equal-width bins and averaged within each bin. The analysis reselects the
maximum decode-minus-prefill mean-shift change over layer, stream, and relative bin under every
task-label permutation. The direct-answer protocol yielded only the first generated-token
update: prefill predicts the answer token, while the next recorded state is conditioned on
that token. Longer trajectories therefore use a separate, matched explanation protocol
with task-specific coverage requirements. The implementation associates updates with each
generated sequence and excludes positions at and after EOS or padding. The reported pilot
uses uncapped batch-size-one capture, for which no second sequence remains active after EOS.
Its first relative bin includes the common answer marker; excluding that marker would
require a separately validated token-boundary rule. The pilot contains 178{,}752 finite
vectors from 32 responses, all with verified residual contributions. Raw responses are
identical with hooks disabled and enabled, update counts range from 25 to 205 (median
47.5), and every response contributes to all eight bins. These checks establish passive
measurement and position coverage for the descriptive pilot, not task-set separation.

\begin{figure}[t]
  \centering
  \includegraphics[width=0.62\textwidth]{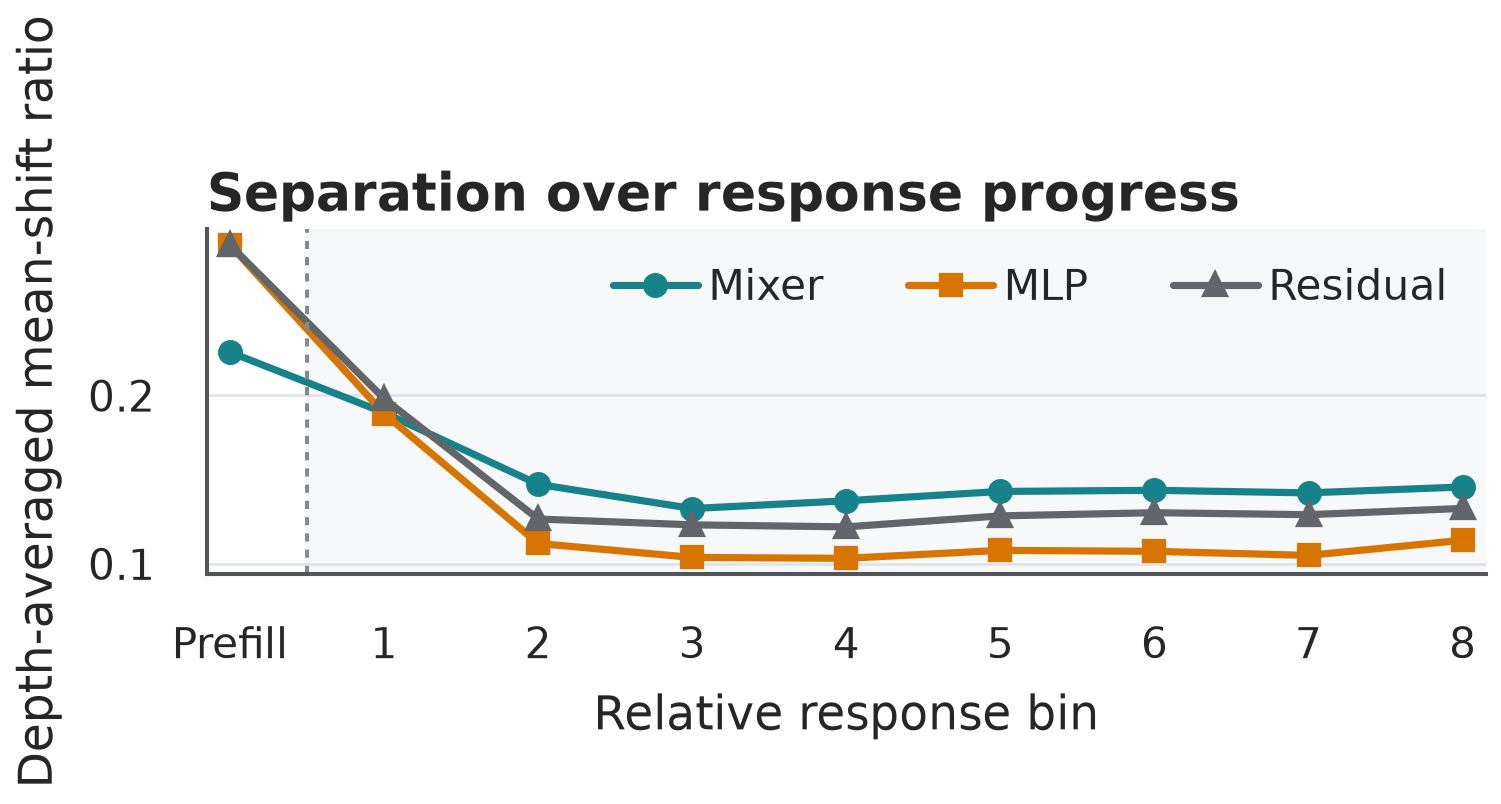}
  \caption{OLMo 3 normalized-response pilot (four tasks per group). The depth-averaged
  mean-shift ratio falls across response bins, although the reselected local maximum (not
  plotted) rises $0.193\!\rightarrow\!0.479$ ($p=0.0286$, the smallest attainable value).
  Effective-rank fractions rise. Bins are fractions of the response length rather than
  semantic reasoning steps.}
  \label{fig:relative-response}
\end{figure}

For the longitudinal study, the OLMo 3 mapping likewise observes the post-attention and
post-feed-forward normalized values added to the residual. The longitudinal source refs and
revision prefixes are
\texttt{stage1-step10000}/\texttt{7717822},
\texttt{stage1-step705000}/\texttt{bc2d50b},
\texttt{stage1-step1413814}/\texttt{373bad2},
\texttt{stage2-step1000}/\texttt{2079c90},
\texttt{stage2-step47684}/\texttt{c3c800d},
\texttt{stage3-step1000}/\texttt{446ae23}, and
\texttt{stage3-step11921}/\texttt{4b32e10}.

\paragraph{Apertus 1.5 endpoint and release scope.}
Apertus 1.5 8B is a suitable second fully open family in principle, and its text decoder
exposes verified token-mixing and feed-forward residual contributions. As of 10 August 2026,
however, the public release provides only its final revision. The model card states that
intermediate checkpoints will follow \citep{apertus15modelcard}. We therefore evaluate
the endpoint implementation but do not report a longitudinal comparison. Apertus 1.0 and
1.5 are not substituted as two trajectory points because the release bundles continued
pretraining, multimodal training, and post-training changes.
The endpoint anchor uses the same ten MMLU subjects as the OLMo trajectory and three
disjoint blocks of 32 examples per task. All 92{,}160 vectors and 288 layer--stream
measurements satisfy the count, finiteness, and verified-contribution criteria.
Knowledge/reasoning accuracy is $0.358\pm0.036/0.306\pm0.022$ across blocks; depth-mean
ratios are $0.204\pm0.016$, $0.241\pm0.018$, and $0.240\pm0.018$ for the token-mixing,
feed-forward, and residual streams. After Benjamini--Hochberg (BH) correction, linear-probe
macro-F1 is significant at $q<0.05$ in all three blocks at 87/96 layer--stream cells. The
blocks measure example-sampling sensitivity, not training-seed variation. Greedy answers,
scores, and same-process forward logits are identical with hooks disabled and enabled. Thus
the endpoint establishes
repeated recoverability and passive extraction, but it supplies no training-time direction.

\paragraph{Apertus 1.0 trajectory validation.}
The separate Apertus 1.0 8B release provides token-indexed intermediate
checkpoints across its 15T-token pretraining run \citep{apertus2025}. Its documented 8B
curriculum uses Stage~1 through 7T tokens, skips Stage~2, and changes the data mixture at
7T, 12T, and 13.5T tokens before the final cooldown. We evaluated 11 immutable checkpoints
spanning the interiors and boundaries of these stages under the same tasks, examples, and
measurement protocol as the OLMo trajectory. The 33 checkpoint--block conditions contain
3{,}168 valid layer--stream measurements with complete counts, finite vectors, verified
residual contributions, fixed revisions, paired samples, and complement-symmetric
statistics, which aggregate to 1{,}056 checkpoint-level layer--stream summaries. The results are reported in
Section~\ref{sec:longitudinal}. The later long-context extension,
SFT/QRPO alignment, and Apertus
1.5 endpoint are excluded from the pretraining curve because they change multiple training
conditions and require separate analyses.
At the 210B, 7.047T, and 15T checkpoints, same-process hook checks cover all 32
decoder layers and 96 verified hook sites. All 384 vectors per checkpoint are finite;
forward logits and greedy token IDs are bit-identical with hooks disabled, enabled, and
removed, and no hooks remain registered after evaluation.

Each evaluation writes a tensor-bearing \texttt{representations.pt}, a vector-free
\texttt{records.csv}, and \texttt{metadata.json}. Offline analysis writes
\texttt{layer\_separability.csv} and \texttt{report.json}. These filenames are an
implementation contract, not part of the scientific contribution. Manifests record
cryptographic hashes, source and package inventories, model and dataset revisions, task/sample identity,
prompts, generation settings, batch policy, precision, hardware, selected model stack,
verified-contribution flags, and failed or skipped measurements.

The MMLU-Pro and heterogeneous-panel study (100 examples per task) contains 100 model--task
conditions and 1{,}320{,}000
vectors. The reported MuSR conditions use the corrected two-choice adapter, and the complete
cross-benchmark panel is evaluated under all 252 balanced task-label permutations. Greedy
outputs match with hooks disabled and enabled on two examples for each evaluated model
configuration; same-process logit equality is also established for the longitudinal OLMo
endpoint. These invariance results apply only to the configurations tested.

The longitudinal extraction contains seven immutable 7B checkpoints and three disjoint
example blocks per checkpoint: 21 conditions and 645{,}120 vectors. Every condition
contains 32 layers, three streams, ten tasks, and 32 examples per task. All 21 conditions
satisfy count, finiteness, verified-contribution, revision, source, paired-ID, and
disjoint-block criteria. The analysis contains 2{,}016/2{,}016 valid
layer--stream--block measurements. It uses the same estimator family and all
252 balanced task-label permutations as the MMLU-Pro analysis.

\paragraph{Inference implementation.}
The reported GPU analysis uses one seeded global task order, filtered under each balanced
task-label permutation. Complement symmetry is verified numerically for the mean-shift and RBF statistics and holds
by construction for the linear probe (Appendix~\ref{app:metric-definitions}).
The eight-vs-eight-subject primary panel has 12{,}870 oriented permutations, including the
observed partition and its complement; equivalently, it has 6{,}435 complement pairs. The
five model conditions enumerate all 12{,}869 non-observed permutations for the primary
mean-shift maximum. The five-vs-five-subject MMLU-Pro analysis analogously uses 252 oriented
permutations, of which 251 are non-observed. Nonlinear inference uses the same examples,
folds, and permutations as its corresponding linear analysis. All reported activation analyses use RAPIDS cuML 26.06 with float64 arithmetic; except for
the 64-component matched comparison, probes operate in a 95\%-variance PCA space fitted on
the training tasks in each fold, and the prompt-only baselines use scikit-learn
(Appendix~\ref{app:analysis-details}). Shared-PCA and raw-space scikit-learn activation
analyses constitute different estimator families and are not combined with these results. Two processes per
model--panel condition run concurrently under NVIDIA MPS.

In the primary panel, the selection-aware mean-shift p-values are $0.0721$, $0.00435$,
$0.00109$, $0.000777$, and $0.0145$ for OLMo, Apertus~1.5, Qwen 9B, Qwen 27B,
and Gemma 4. Across layer--stream cells, BH-adjusted significance occurs in 90/96,
95/96, 95/96, 191/192, and 150/180 cells, respectively. The exact p-values for these
cell-count statistics are $0.00155$, $0.00684$, $0.00311$, $0.00124$, and $0.00824$.
Because layer--stream cells are dependent, these counts are diagnostics rather than
model-level effect sizes.

The selected-site linear-probe analysis performs 6{,}435 probe refits per model, one for
each label-complement pair, while preserving all 12{,}870 oriented task-label permutation scores.
Accuracy and macro-F1 are complement symmetric by construction in all five reports. The macro-F1
randomization p-values are $0.00155$, $0.00186$, $0.00186$, $0.00420$, and $0.00404$ in the
model order above. This is an exact test of the prespecified composite statistic, not a causal
or prompt-adjusted result.

Repeated RBF fits do not reproduce all scores exactly, so threshold-adjacent RBF results are
treated as numerically unstable rather than as evidence of a qualitative linear--nonlinear
contrast. Silhouette scores are computed from explicit batched cosine distances.

\section{Experimental Details}
\label{app:experimental-details}

\subsection{Cross-model configurations}

\begin{table}[h]
\centering
\small
\caption{Evaluated conditions on one 94.5-GiB GH200. All use deterministic prefill
extraction, final-valid-token pooling, and 100 examples per task. Batch is the extraction
batch size, not the number of analyzed examples.}
\label{tab:configurations}
\begin{tabular}{llllr}
\toprule
Model & Revision & Layers & Storage & Batch \\
\midrule
OLMo 3 7B Instruct & \texttt{6e5971d} & 32 & float32 & 32 \\
Apertus 1.5 8B & \texttt{a411d83} & 32 & float32 & 24 \\
Qwen 3.5 9B & \texttt{c202236} & 32 & float32 & 20 \\
Qwen 3.5 27B & \texttt{fc05dae} & 64 & float32 & 8 \\
Gemma 4 31B IT & \texttt{842da37} & 60 & float32 & 2 \\
\bottomrule
\end{tabular}
\end{table}

\subsection{Assets and use terms}

We download public model and benchmark releases at the recorded revisions and do not
redistribute their weights or benchmark text. The release metadata identify OLMo~3,
Apertus~1.0/1.5, Qwen~3.5, and Gemma~4 as Apache-2.0 releases. MMLU and MMLU-Pro are MIT;
GSM8K is MIT; ARC is CC-BY-SA-4.0; Natural Questions Open is CC-BY-SA-3.0; MuSR is
CC-BY-4.0; HellaSwag is MIT; and SuperGPQA is ODC-BY with additional attribution duties
for incorporated sources. The mirrored TriviaQA and MedQA dataset cards do not state a
license. We use them only through the public evaluation interface, do not redistribute
their content, and do not claim that public availability supplies redistribution rights.
The implementation builds on MIT-licensed \texttt{lmms-eval}, Apache-2.0 Transformers and
RAPIDS, and BSD-licensed PyTorch and scikit-learn. These statements summarize the source
metadata; redistribution remains subject to the applicable licenses, notices, and Apertus
acceptable-use terms.

\subsection{Measured compute}

All reported extraction and GPU analyses ran on one NVIDIA GH200 with 94.5~GiB usable
memory. Telemetry spans 1.24 hours for the five-model eight-vs-eight-subject extraction, 6.79 hours for its
selection-aware probe, and 2.31 hours for the 21-condition OLMo analysis. The Apertus 1.0
analysis used 23.7 logged GPU-hours, including interrupted computation; this is not a
clean-run timing estimate. These are single-device wall-clock
intervals, not normalized training FLOPs. The matched prompt comparison, lexical diagnostics,
and fixed-site calibration share a 3.00-hour telemetry trace that includes unsuccessful
solver computation and therefore is not a clean-run timing estimate. Additional exploratory
and validation computation is excluded from these totals.

\subsection{Text-only cross-model panels}
\label{app:text-extension}

Before model execution, the protocol records the exact 100 document indices and
processed-dataset fingerprint for every task; validation requires exact agreement in every
model condition. One panel uses a
common MMLU-Pro prompt and answer format: biology, health, history, law, and economics are
knowledge-oriented proxies; mathematics, physics, computer science, engineering, and
philosophy are reasoning-oriented proxies. MMLU-Pro increases reasoning demand relative to
MMLU, but its subject categories still mix both demands \citep{wang2024mmlupro}.

A second, deliberately heterogeneous stress test pairs closed-book Natural Questions and
TriviaQA \citep{kwiatkowski2019naturalquestions,joshi2017triviaqa} with MedQA, SuperGPQA,
and MMLU-Pro history on the knowledge-oriented side, and GSM8K, MuSR murder mysteries,
ARC-Challenge, HellaSwag, and MMLU-Pro mathematics on the reasoning-oriented side
\citep{cobbe2021gsm8k,sprague2024musr,clark2018arc,zellers2019hellaswag,du2025supergpqa}.
This taxonomy follows benchmark objectives only comparatively: GSM8K and MuSR explicitly
target multistep reasoning; no-context Natural Questions and TriviaQA most directly target
closed-book retrieval; ARC and SuperGPQA require both knowledge and reasoning; HellaSwag is
commonsense continuation inference rather than formal reasoning. The panel therefore tests
whole-task transfer under broader task sampling, not construct validity. For MuSR, the
evaluated adapter renders exactly two choices for all 500 model--example records;
string-encoded answer choices are not split into individual characters.

The evaluated current-generation set is OLMo 3 7B Instruct \citep{olmo3}, Apertus~1.5 8B
\citep{apertus2025,apertus15modelcard}, Qwen 3.5 9B/27B \citep{qwen35}, and Gemma 4 31B
\citep{gemma4}. We exclude model configurations whose managed reasoning traces are incompatible with
the fixed short-answer protocol rather than changing generation rules by model.

Prompt token counts differ between the task sets despite the common text modality. In the MMLU-Pro five-vs-five-subject panel, mean knowledge-task counts span $274$--$308$
tokens and mean reasoning-task counts $229$--$267$, depending on tokenizer. A one-feature logistic probe, with the same task-held-out folds and
all 252 task-label permutations, does not predict that panel's label significantly (macro-F1
randomization $p=0.167$--$0.270$). These values do not describe the primary MMLU panel.

In the MMLU eight-vs-eight-subject panel, the corresponding means are $179.0/135.3$ for OLMo, $172.0/127.1$ for both
Qwen sizes, $172.2/129.5$ for Gemma, and $206.6/163.5$ for Apertus. Prompt length is
therefore strongly associated with the prespecified partition even though it was not sufficient
by itself in the MMLU-Pro panel. The exact TF--IDF control in
Section~\ref{sec:cross-model-results} uses the logged task
prompt before model-specific chat-template rendering. Its vocabulary, inverse-document
frequencies, and classifier are refit within every held-out-task fold and balanced
task-label permutation. Significance under this null therefore shows that task content
already predicts the prespecified partition; it does not quantify all surface information in the model's rendered input or
establish that lexical variation causes the representation result.

The primary MMLU evaluation records retain parsed outputs but not raw generated text when the
parser fails. Valid A--D outputs occur in $0.992$, $0.573$, $1.000$, $1.000$, and $0.761$ of
examples for OLMo, Apertus~1.5, Qwen 9B, Qwen 27B, and Gemma, respectively. The primary
representations are measured before generation and remain usable; cross-model behavioral
accuracy and failure-conditioned analyses are not interpreted. Task-level validity spans
$0.93$--$1.00$ for OLMo, $0.20$--$0.83$ for Apertus~1.5, $1.00$ for both Qwen sizes,
and $0.16$--$1.00$ for Gemma; the released result tables list every model--task condition.

The primary extraction uses the final valid prompt token during prefill and float32
storage. A descriptive decoding pilot is reported above. Mean-valid-token pooling has not
yet been evaluated on the complete panel and is discussed as a limitation.
Batch sizes are fixed within model after peak-memory calibration. Representations collected
under different batch sizes are analyzed separately because calibration showed small but
nonzero changes despite identical generated responses.

\subsection{Supplementary multimodal benchmark analysis}
\label{app:multimodal-side}

We conduct a separate cross-benchmark multimodal analysis with Qwen 3.5 9B/27B and Gemma 4
31B. We restrict the comparison to these current-generation multimodal checkpoints and
exclude older Gemma 3 checkpoints to match the model-generation rule used in the primary
analysis. This analysis is supplementary and does not alter the text-only primary comparison. The prespecified
knowledge-oriented group contains OK-VQA, MedQA, MMMU, AI2D, and GPQA
\citep{marino2019okvqa,jin2021medqa,yue2024mmmu,kembhavi2016ai2d,rein2024gpqa}; the
capability-oriented group contains POPE, MME, TextVQA, OCRBench v2, and DocVQA
\citep{li2023pope,fu2025mme,singh2019textvqa,fu2025ocrbenchv2,mathew2021docvqa}.
These operational groups are not matched constructs. All five capability-oriented tasks use
images, compared with three of five knowledge-oriented tasks, and the groups also differ in
domain, benchmark family, prompt content, and response format.

We retain 100 examples per task and the same final-prompt-token, three-stream extraction.
All 30 model--task conditions have complete counts, finite vectors, verified residual
contributions, and valid task prompts. Inference holds out complete tasks and enumerates 252
oriented task-label permutations, or 126 label-complement pairs, so the minimum
class-symmetric p-value is $1/126=0.00794$. The mean-shift maximum is recomputed under every
permutation; each selected-site probe then uses that permutation-specific site. Controls
remove, within each training-task fold, activation
variance linearly predicted from six prompt-surface variables, task-level image presence,
three response-format indicators, a 64-component character TF--IDF basis, or both sets.
Image presence does not encode pixels, visual-token count, or image content.

\begin{table}[h]
\centering
\scriptsize
\setlength{\tabcolsep}{3.2pt}
\caption{Supplementary heterogeneous multimodal analysis. Mean-shift cells report the
selection-aware maximum, zero-indexed layer and stream (T: token-mixing update; R: residual state), and
exact p-value. Probe cells report held-out-task macro-F1 and exact p-value in parentheses.
No result is significant under its exact task-label null at $p<0.01$.}
\label{tab:multimodal-side}
\begin{tabular}{lccccc}
\toprule
Model & Mean-shift maximum & Raw probe & Surface+modality & Lexical-64 & Combined \\
\midrule
Qwen 3.5 9B  & $1.057$, L0-T ($0.0238$)  & $0.879$ ($0.0159$) & $0.210$ ($0.960$) & $0.591$ ($0.349$) & $0.151$ ($0.992$) \\
Qwen 3.5 27B & $1.199$, L5-T ($0.0159$)  & $0.858$ ($0.0317$) & $0.415$ ($0.698$) & $0.550$ ($0.373$) & $0.310$ ($0.952$) \\
Gemma 4 31B  & $1.113$, L36-R ($0.0476$) & $0.899$ ($0.0159$) & $0.538$ ($0.421$) & $0.735$ ($0.0635$) & $0.336$ ($0.913$) \\
\bottomrule
\end{tabular}
\end{table}

\begin{table}[H]
\centering
\scriptsize
\caption{Complementary measurements at the observed mean-shift-selected site. These quantities
were not given task-randomization tests in this supplementary analysis and are descriptive.
Paired CKA is undefined, rather than zero, because the two benchmark groups have no
declared example pairing; covariance alignment supplies an order-invariant comparison.}
\label{tab:multimodal-side-descriptive}
\begin{tabularx}{\textwidth}{@{}lcccc>{\raggedright\arraybackslash}X@{}}
\toprule
Measurement at selected site & Qwen 9B & Qwen 27B & Gemma 31B & Reference & Role \\
\midrule
RBF probe macro-F1 & $0.837$ & $0.850$ & $0.799$ & $0.50$ & Nonlinear prediction; descriptive here \\
Cosine silhouette & $0.282$ & $0.256$ & $0.434$ & $0$ & Higher means greater distance-based separation \\
Covariance alignment & $0.421$ & $0.375$ & $0.080$ & $1$ & Lower means less similar unpaired geometry \\
Within-group effective-rank fraction & $0.109$ & $0.326$ & $0.418$ & $1$ & Lower means a more concentrated spectrum \\
Within-group stable rank & $1.202$ & $1.510$ & $2.701$ & --- & Scale-dependent spectral concentration diagnostic \\
Paired linear CKA & --- & --- & --- & $1$ & Undefined: zero declared cross-group pairs \\
\bottomrule
\end{tabularx}
\end{table}

The raw macro-F1 values exceed the $0.781$--$0.805$ range in the primary MMLU panel, consistent
with additional modality and response-format cues that predict the group label; this is not
evidence of stronger construct separation. Their exact p-values are $0.0159$--$0.0317$, and prompt-only
word and character TF--IDF baselines reach macro-F1 $0.804$ and $0.802$ (both $p=0.0159$).
Adjustment sharply reduces the representation scores (Table~\ref{tab:multimodal-side});
Table~\ref{tab:multimodal-side-descriptive} lists the complementary measurements at the
selected sites.
This establishes sensitivity to measured task and prompt differences; it neither identifies
a modality-independent knowledge/reasoning signature nor proves that the measured
variables causally explain the raw score. The result therefore remains separate from the
primary text-only evidence.

\section{Metric Definitions and Inference}
\label{app:metric-definitions}

\begin{figure}[t]
\centering
\includegraphics[width=\textwidth]{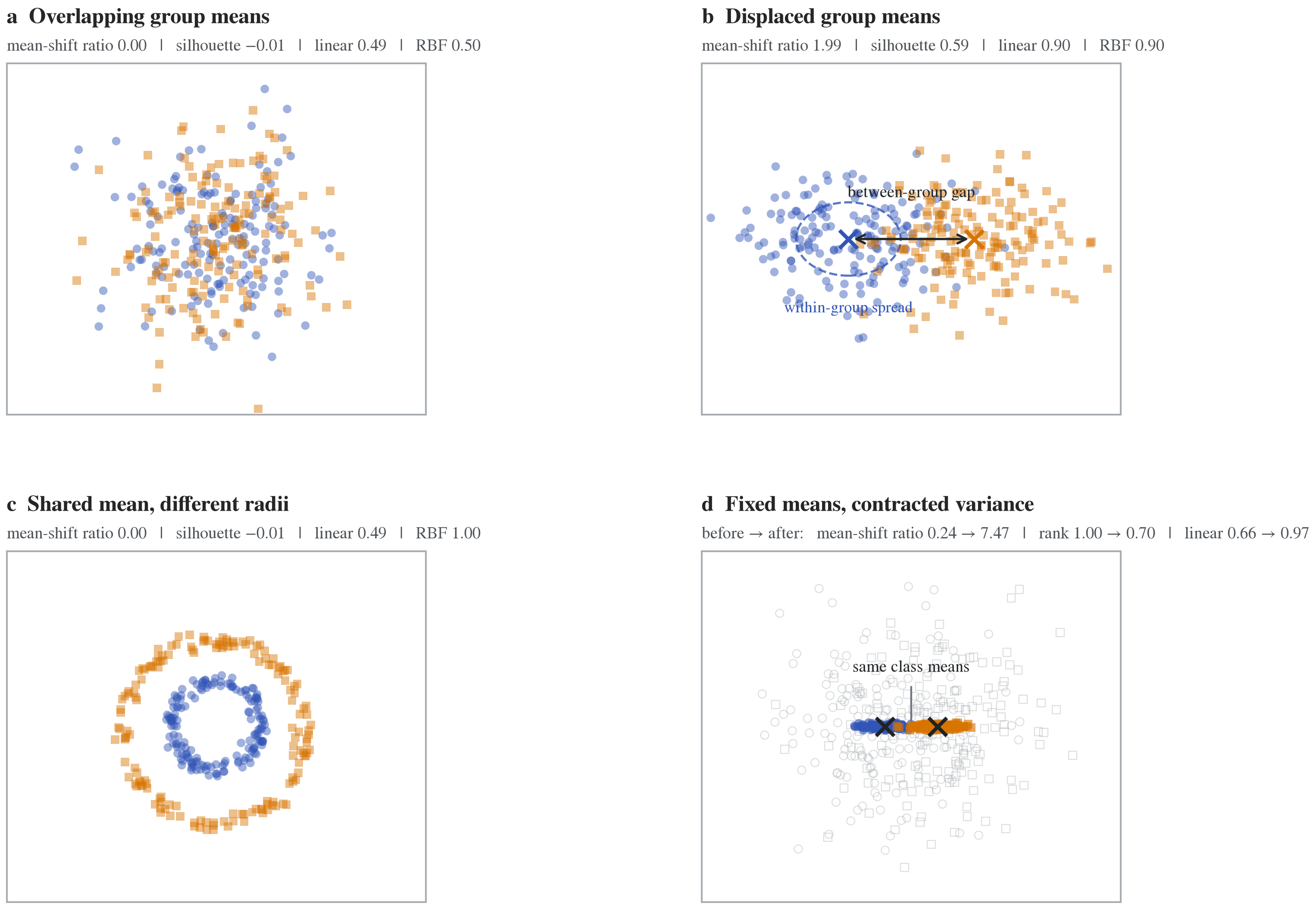}
\caption{Synthetic metric guide computed directly in two dimensions; no embedding method is
used. Blue circles denote $K$, orange squares denote $R$, crosses mark class means, and gray
hollow points in (d) show the state before contraction. Displaced means raise the mean-shift ratio and
both probes (b), whereas equal-mean concentric groups are detected only by the nonlinear
probe (c). Holding the means fixed while reducing within-group variance sharply raises
the mean-shift ratio (d), illustrating why we report its numerator, denominator, and effective rank
separately. Probe values use five complete-group folds with fold-local preprocessing. The
examples illustrate what each statistic responds to and define no empirical cutoffs.}
\label{fig:metric-illustration}
\end{figure}

Table~\ref{tab:metric-references} summarizes the four measurement families and their
reference values, and Figure~\ref{fig:metric-illustration} illustrates them on synthetic data.

\begin{table}[h]
\centering
\footnotesize
\caption{Four complementary measurement families. Reference values aid interpretation;
they are not universal quality thresholds. Colors are redundant with the family labels.}
\label{tab:metric-references}
\begin{tabularx}{\textwidth}{@{}
  >{\raggedright\arraybackslash}p{0.18\linewidth}
  >{\raggedright\arraybackslash}p{0.29\linewidth}
  >{\raggedright\arraybackslash}X@{}}
\toprule
Family & Reference value & Scientific role \\
\midrule
\fishermetric & $0$: equal projected means; $1$: squared mean shift equals summed within-category variance &
Within-model mean separation; numerator and denominator remain visible. \\
\probemetric & $0.50$: balanced chance &
Prediction for tasks excluded from probe fitting; task-label permutations define the null. \\
\geometrymetric & Silhouette $0$: no distance advantage; alignment $1$: identical normalized geometry &
Cohesion and similarity; paired CKA is reported only for actual pairs. \\
\rankmetric & $1$: maximally even spectrum &
Contraction diagnostic, not separation evidence by itself. \\
\bottomrule
\end{tabularx}
\end{table}

\paragraph{PCA-space mean-shift ratio.}
For each model, zero-indexed layer $\ell\in\{0,\ldots,L-1\}$, and stream $s$, PCA retains the smallest basis explaining 95\% of
variance. With projected category means $\mu_K,\mu_R$ and component variances
$\sigma^2_{K,j},\sigma^2_{R,j}$, the primary descriptive statistic is
\[
  R_{\ell,s}=
  \frac{\sum_j(\mu_{K,j}-\mu_{R,j})^2}
       {\sum_j(\sigma^2_{K,j}+\sigma^2_{R,j})+10^{-12}}.
\]
We also retain per-component ratios, the numerator, the denominator, retained
dimension, and explained variance. A value of zero means that the projected group means
coincide; a value of one means that their squared displacement equals the summed
within-group variance. The value is not a
classification probability and has no model-independent cutoff. Unlike classical Fisher
linear discriminant analysis, the denominator is the summed within-category variance
$\sum_j(\sigma^2_{K,j}+\sigma^2_{R,j})$ rather than a regularized inverse-covariance
quadratic form. It therefore describes total normalized mean
shift but need not rank sites by linear separability. The analysis files retain the legacy
field \texttt{FDR\_trace}; significance is calibrated by task-label permutation rather than
by the raw magnitude.

\paragraph{Rank diagnostics.}
For centered activation matrix $A$ with singular values $\sigma_k$, define
\[
  p_k=\frac{\sigma_k}{\sum_j\sigma_j},\qquad
  \operatorname{erank}(A)=
  \exp\left(-\sum_{k:p_k>0}p_k\log p_k\right).
\]
We compute effective and stable rank for all examples and after subtracting the corresponding
knowledge- or reasoning-category mean. Effective rank is never used as a divisor: dividing by a falling rank would enlarge
the adjusted mean-shift value and amplify the pattern it is intended to diagnose.

\paragraph{Probes and finite task-level permutation tests.}
The linear probe uses fold-local PCA, fold-local standardization, and balanced logistic
regression. The nonlinear probe uses the same preprocessing and a balanced RBF classifier.
Five-fold cross-validation holds out complete tasks from both categories, and pooled
out-of-fold predictions yield accuracy and macro-F1. For probe score $T$, labels are
permuted across tasks while preserving the number of tasks per category, and the complete
probe is refit:
\[
  p=\frac{1+\sum_{b=1}^{B}\mathbf{1}[T_b\geq T_{\mathrm{obs}}]}{B+1}.
\]
Five tasks per category yield 252 oriented balanced task-label permutations, including the observed
partition. Accuracy, macro-F1, and the PCA-space mean-shift ratio are invariant to exchanging
the two class names, so these permutations form 126 complementary equivalence classes.
Enumerating all 251 alternatives
therefore gives minimum attainable $p=2/252=1/126$ for these class-symmetric statistics
\citep{ojala2010permutation}. Mean-shift and RBF label-complement orientations are evaluated
independently. For the linear probe, one fitted
orientation is reused for its complement because accuracy and macro-F1 are mathematically
invariant to exchanging class names; this equality is imposed rather than checked
numerically. Mean-shift-selected cell identity is checked independently. A
symmetry failure in an independently evaluated statistic makes its exact inference
undefined. Benjamini--Hochberg (BH) correction at
$0.05$ consequently requires at least 16 minimum-$p$ cells when $m=96$, 29 when $m=180$, and
31 when $m=192$ before any cell can be significant after correction.
For mean-shift inference, PCA is label-blind and fixed within a layer--stream cell. We aggregate
sample counts, activation sums, and squared sums by task, recompute the PCA-space mean-shift ratio for
every permutation, and reselect the maximum across all prespecified cells. Each permutation is
also treated as observed in turn to obtain its cellwise p-vector and BH-significant-cell count;
the observed count is then compared with this finite distribution. The aligned null
scores are stored in a compressed analysis file.
Benjamini--Hochberg correction is applied separately to accuracy and macro-F1 across the
prespecified layer--stream family \citep{benjamini1995fdr}. Undefined cells retain missing values but still
count toward the prespecified family size. Nonlinear probe scores are
descriptive unless their own permutation refits are requested.

\paragraph{Geometry.}
For explicitly paired centered matrices $X,Y$, linear CKA is
\[
  \mathrm{CKA}(X,Y)=
  \frac{\norm{X^\top Y}_F^2}
       {\norm{X^\top X}_F\norm{Y^\top Y}_F}.
\]
It is undefined for unrelated benchmark examples. For unpaired sets we instead report
normalized covariance alignment,
\[
  \rho_{\mathrm{cov}}=
  \frac{\langle C_K,C_R\rangle_F}
       {\norm{C_K}_F\norm{C_R}_F}.
\]
Cosine silhouette uses $s_i=(b_i-a_i)/\max(a_i,b_i)$, where $a_i$ is mean within-group
distance and $b_i$ is distance to the other group \citep{rousseeuw1987silhouettes}.
A zero silhouette means equal average within- and other-group distance; positive values
place an example closer to its assigned group on average. We do not import fixed clustering
quality cutoffs into this supervised, high-dimensional setting. Silhouette and covariance
are computed in the original activation space; CKA is reported only when explicit pairs
exist.

\section{Analysis Hyperparameters and Reporting Fields}
\label{app:analysis-details}

The reported linear estimator uses fold-local GPU PCA and standardization followed by class-balanced
logistic regression with $C=1$, $\ell_2$ regularization, the cuML quasi-Newton solver, and
a 3000-iteration ceiling. The implementation enables cuML's penalty normalization. This is
the frozen estimator definition, not an assertion of equivalence to scikit-learn's $C=1$:
the libraries scale their objectives differently and the effective penalty can depend on
the training-set size. The nonlinear estimator is a class-balanced GPU RBF SVM with
$C=1$ and scale-adaptive bandwidth. Both probes operate in a 95\%-variance PCA space fit
only on training tasks in each fold. The descriptive PCA-space mean-shift ratio uses a separate
global 95\%-variance projection for each layer and stream. The silhouette calculation uses
explicit batched cosine distance; the primary eight-vs-eight-subject panel is evaluated exactly, while larger
conditions use seeded class-balanced subsampling above the prespecified cap. The primary
eight-vs-eight-subject analysis enumerates all 12,870 oriented balanced task-label
permutations, comprising the observed partition and 12,869 alternatives. Class-symmetric
statistics form 6,435 complementary pairs, including the observed pair; thus 6,434 pairs
belong to the null relative to that pair. The MMLU-Pro and heterogeneous-panel linear analyses enumerate all 252 permutations for
five tasks per category, and the cross-model nonlinear
sensitivity analysis uses the same smaller permutation set; the
longitudinal RBF scores remain descriptive because that study did not request nonlinear
permutations.

The prompt-only baselines use scikit-learn balanced logistic regression with the liblinear
solver, $C=1$, and $\ell_2$ regularization. Word TF--IDF uses lowercased 1--2-grams;
character TF--IDF uses lowercased within-word 3--5-grams and at most 100,000 features. The
vocabulary, inverse-document frequencies, and classifier are refit inside each task-held-out
fold and balanced task-label permutation. These baselines share the inferential unit and permutation set
with the representation probe, but not its feature space or GPU estimator.

The prompt-feature sensitivity standardizes its covariates on the training tasks and fits a
multi-output Ridge map ($\alpha=1$) from those covariates to every activation dimension.
Training and held-out activations are residualized before fold-local PCA and probing. The
map is refit for every fold and task-label permutation. The length-only condition uses the
model-tokenizer prompt length; the six-feature condition additionally uses character and
line counts, option count, whitespace-token count, and whitespace-token entropy. This
operation removes linearly predictable variance, including any construct-correlated
variance carried by those covariates; it is a sensitivity analysis, not causal adjustment.

The lexical sensitivity uses the same character TF--IDF definition as the prompt baseline.
For every task-label permutation, the unadjusted PCA-space mean-shift profile is used to reselect the
layer--stream site. Within each resulting probe fold, the TF--IDF vocabulary and inverse
document frequencies are fit on training-task prompts, randomized truncated SVD maps them
to 64 components, and a multi-output Ridge map predicts the activation coordinates. The
residual activations then enter fold-local PCA, standardization, and probing. A 32-component
run tests dimensionality sensitivity. These logged prompts precede model-specific chat
rendering, and a finite low-rank linear map does not exhaust lexical information.

The matched excess test places representations and prompt text under one estimator family.
For every balanced task-label permutation, it reselects the unadjusted mean-shift site, maps both feature sets
to 64 fold-local components, standardizes them, and fits the same cuML logistic probe with
the same grouped folds and regularization. Its statistic is the paired difference in
macro-F1, $\Delta=\operatorname{macro\text{-}F1}_{\mathrm{representation}}-
\operatorname{macro\text{-}F1}_{\mathrm{prompt}}$; all 12,870 permutation-specific differences form the exact
one-sided reference distribution. Equal feature count and classifier do not make activation
and TF--IDF feature spaces semantically equivalent, so a positive $\Delta$ remains an
incremental prediction result rather than a causal claim.

\begin{table}[h]
\centering
\footnotesize
\setlength{\tabcolsep}{3.5pt}
\caption{Matched activation-minus-prompt inference and width sensitivity. The null column
summarizes 6,434 non-observed label-complement pairs. PCA-95 replaces activation PCA-64
while retaining 64 prompt components; it tests compression rather than capacity matching.
Every observed $\Delta$ lies within its null distribution.}
\label{tab:review-validation}
\begin{tabular}{lccc}
\toprule
Model & Matched-64 $\Delta$/exact $p$ & Null mean $\pm$ SD; q95 & PCA-95 $\Delta$/exact $p$ \\
\midrule
OLMo 3 7B & $0.048/0.290$ & $0.016\pm0.057; 0.108$ & $0.025/0.466$ \\
Apertus 1.5 8B & $0.070/0.177$ & $0.011\pm0.064; 0.112$ & $0.049/0.257$ \\
Qwen 3.5 9B & $0.047/0.291$ & $0.011\pm0.066; 0.123$ & $0.050/0.258$ \\
Qwen 3.5 27B & $0.017/0.440$ & $0.006\pm0.067; 0.112$ & $0.041/0.295$ \\
Gemma 4 31B & $0.051/0.297$ & $0.017\pm0.066; 0.129$ & $0.050/0.330$ \\
\bottomrule
\end{tabular}
\end{table}

Table~\ref{tab:review-validation} reports two additional analyses of this comparison. Null
summaries use one value from each of the 6,434 null label-complement pairs; the exact
test retains all 12,870 oriented task-label permutations. Replacing activation PCA-64 by each model's
fold-local 95\%-variance activation basis changes neither the sign nor the inference. This
unequal-width comparison tests compression sensitivity, not matched capacity.

We also replace the PCA-space mean-shift selector with the regularized whitened mean displacement
$\Delta\mu^\top(W+\lambda I)^{-1}\Delta\mu$ in the same label-blind 95\%-variance
projection. Here $W$ is the permutation-specific pooled within-category covariance and
$\lambda=\alpha\,\operatorname{tr}(C)/d$, where $C$ is the label-blind total covariance and
$d$ its dimension.
For $\alpha\in\{0.01,0.1,1\}$, site selection is repeated for every balanced task-label permutation.
The vectorized calculation agrees with direct covariance inversion within
$2.4\times10^{-13}$ on sampled permutations. The exact test rejects the task-label null for every
maximum after Holm correction over the 15-condition grid
($p_{\mathrm{Holm}}\leq0.0094$), but none selects the PCA-space mean-shift site
(Table~\ref{tab:selector-sensitivity}).
Thus recoverability is not specific to the PCA-space mean-shift ratio, whereas layer--stream localization
depends strongly on the selector and regularization.

\begin{table}[t]
\centering
\small
\setlength{\tabcolsep}{4pt}
\caption{Selector sensitivity on the primary panel. Each entry gives the zero-indexed layer and
stream selected by the observed partition; the exact test reselects over all layers and
streams for each of 12,870 task-label permutations. These sites are neither causal claims nor a
cross-panel transfer test.}
\label{tab:selector-sensitivity}
\resizebox{\textwidth}{!}{%
\begin{tabular}{lccccc}
\toprule
Model & Mean-shift site & $\alpha=0.01$ & $\alpha=0.1$ & $\alpha=1$ & Exact $p$ range \\
\midrule
OLMo 3 7B & 3 / feed-forward & 21 / residual & 20 / residual & 6 / residual & $0.00062$--$0.00078$ \\
Apertus 1.5 8B & 0 / feed-forward & 18 / residual & 4 / residual & 4 / residual & $0.00062$--$0.00109$ \\
Qwen 3.5 9B & 2 / feed-forward & 30 / residual & 30 / residual & 30 / residual & $0.00093$--$0.00233$ \\
Qwen 3.5 27B & 1 / token mixing & 25 / token mixing & 25 / token mixing & 25 / token mixing & $0.00062$--$0.00218$ \\
Gemma 4 31B & 23 / feed-forward & 31 / token mixing & 31 / token mixing & 22 / token mixing & $0.00078$--$0.00124$ \\
\bottomrule
\end{tabular}
}
\end{table}

The inferential-unit diagnostic uses the alphabetical eight-vs-eight-subject partition and its corresponding
layer-0 feed-forward site in OLMo. Random example folds place every task in both train and
test and give macro-F1 $0.802$ ($p=0.0001$ over 9,999 sample-label permutations). Holding out
complete tasks gives $0.574$ ($p=0.228$ over all balanced task-label permutations). The former test
is deliberately invalid for a benchmark-category claim; it shows empirically why examples
cannot be treated as exchangeable task replicates. It implements the validation contrast,
not a reproduction of Sahoo et al.'s datasets, model, or intervention.

Detection calibration uses the observed OLMo selected site and an arbitrary balanced split
of the fixed task order, not the knowledge/reasoning labels. Before lexical adjustment, we
add a seeded unit-direction shift with standardized mean differences $0$, $0.5$, $1$, $2$,
and $4$. Character TF--IDF fitting, Ridge adjustment, 95\%-variance PCA, and the linear probe
are then repeated within the grouped folds for every balanced task-label permutation. Ten direction
seeds are evaluated at each nonzero shift. This identifies the signal scale detectable by
the adjusted fixed-site instrument in one model; it does not calibrate layer selection or
measure a model mechanism.

\begin{table}[t]
\centering
\small
\setlength{\tabcolsep}{7pt}
\caption{Fixed-site detection calibration after fold-local lexical adjustment. The injected
mean difference is measured in standard deviations of the unadjusted activation projected
onto each seeded unit direction. Every condition uses the complete 12,870-permutation null;
the zero-shift row is evaluated once because direction is immaterial. Direction seeds are
calibration conditions, not model, data, or training replicates.}
\label{tab:probe-calibration}
\begin{tabular}{rrrr}
\toprule
Standardized mean difference & Directions & Exact $p<0.01$ & Macro-F1 mean [range] \\
\midrule
$0$   & 1  & 0 & $0.472$ \\
$0.5$ & 10 & 0 & $0.475$ $[0.471,0.479]$ \\
$1$   & 10 & 0 & $0.483$ $[0.476,0.487]$ \\
$2$   & 10 & 0 & $0.513$ $[0.493,0.528]$ \\
$4$   & 10 & 5 & $0.724$ $[0.638,0.815]$ \\
\bottomrule
\end{tabular}
\end{table}

A separate selection-aware calibration uses the alphabetical eight-versus-eight-task
partition in OLMo. It appends an injected candidate site to the real-site mean-shift matrix,
then reselects the site and refits the matched activation-minus-prompt statistic for every
task-label permutation. Across ten seeded directions at each standardized difference
$0.5$, $1$, $2$, and $4$, the synthetic site is selected for the alphabetical partition
0/10 times and the exact test detects 0/10 directions. This result calibrates the interaction
between the unwhitened selector and this particular injected site; it does not imply that no
selection-aware procedure can detect a shift of that magnitude.

Grouped folds are rebuilt from the two task classes for every task-label permutation. Across all
12,870 permutations, each of the first three test folds contains two tasks per class and
each of the remaining folds one per class; no permuted fold is single-class or less balanced
than the observed fold.

For reproducibility, each condition is identified by model and dataset
revision, document IDs, task split, prompt, sample count, generation settings, numerical
precision, backend, and batching policy. Prompt and output lengths, correctness, semantic
stream mappings, skipped records, and analysis failures are retained for sensitivity analyses and
auditing. Undefined quantities, including unpaired CKA, are reported as missing rather than set
to zero.

\paragraph{Uncertainty sources.}
We keep four sources distinct. Exact task-label permutations quantify evidence against the
probe null. Disjoint example blocks measure sensitivity to selected examples. Repeating
fold seeds measures analysis-split sensitivity. Independently trained checkpoints are
required for training-seed uncertainty. The current longitudinal table has the first two
but only one released training trajectory per model; its mean and standard deviation must not be
presented as a confidence interval over model training. Greedy extraction on identical
examples is deterministic, so repeating generation seeds would add duplicate observations.

\section{Use of Large Language Models}

General-purpose language-model tools assisted with codebase auditing, implementation
review, and editorial revision. The authors remain responsible for the study design,
source verification, code, experiments, analysis, and all claims.

\ifdefined\arxivversion\else
\clearpage
\section*{NeurIPS Paper Checklist}

\begin{enumerate}

\item {\bf Claims}
    \item[] Question: Do the main claims made in the abstract and introduction accurately reflect the paper's contributions and scope?
    \item[] Answer: \answerYes{}
    \item[] Justification: The abstract, Introduction, Results, Limitations, and Conclusion state the observational scope and distinguish task-set predictability from a general mechanism or causal claim.

\item {\bf Limitations}
    \item[] Question: Does the paper discuss the limitations of the work performed by the authors?
    \item[] Answer: \answerYes{}
    \item[] Justification: The Limitations and Broader Impact section discusses construct confounding, limited task-level power, one training trajectory, pooling, observer overhead, and the boundary between passive and causal evidence.

\item {\bf Theory assumptions and proofs}
    \item[] Question: For each theoretical result, does the paper provide the full set of assumptions and a complete (and correct) proof?
    \item[] Answer: \answerNA{}
    \item[] Justification: The paper reports an empirical measurement study and makes no theorem or formal proof claim.

    \item {\bf Experimental result reproducibility}
    \item[] Question: Does the paper fully disclose all the information needed to reproduce the main experimental results of the paper to the extent that it affects the main claims and/or conclusions of the paper (regardless of whether the code and data are provided or not)?
    \item[] Answer: \answerYes{}
    \item[] Justification: The Experimental Details, Implementation and Verification Details, Metric Definitions, and Analysis Hyperparameters appendices report model and data revisions, task composition, extraction settings, estimators, and inference procedures.

\item {\bf Open access to data and code}
    \item[] Question: Does the paper provide open access to the data and code, with sufficient instructions to faithfully reproduce the main experimental results, as described in supplemental material?
    \item[] Answer: \answerNo{}
    \item[] Justification: The implementation and aggregate artifacts are currently in private repositories. The paper gives detailed reproduction information, but no anonymous public code package accompanies this draft yet.

\item {\bf Experimental setting/details}
    \item[] Question: Does the paper specify all the training and test details (e.g., data splits, hyperparameters, how they were chosen, type of optimizer) necessary to understand the results?
    \item[] Answer: \answerYes{}
    \item[] Justification: Sections 4--7 and the experimental appendices specify the frozen checkpoints, tasks, sample blocks, prompts, decoding, pooling, precision, batch sizes, probe settings, and multiplicity correction.

\item {\bf Experiment statistical significance}
    \item[] Question: Does the paper report error bars suitably and correctly defined or other appropriate information about the statistical significance of the experiments?
    \item[] Answer: \answerYes{}
    \item[] Justification: The paper reports sample standard deviations across three paired example blocks, identifies what those repetitions do and do not measure, and uses exhaustive task-label randomization with complement-symmetry checks and Benjamini--Hochberg correction.

\item {\bf Experiments compute resources}
\item[] Question: For each experiment, does the paper provide sufficient information on the computer resources (type of compute workers, memory, time of execution) needed to reproduce the experiments?
    \item[] Answer: \answerNo{}
    \item[] Justification: The appendix identifies the GPU, memory, process layout, batch sizes, and measured wall-clock intervals for the largest extraction and analysis jobs. A complete project-wide total, including failed and preliminary runs, has not yet been consolidated.
    
\item {\bf Code of ethics}
    \item[] Question: Does the research conducted in the paper conform, in every respect, with the NeurIPS Code of Ethics \url{https://neurips.cc/public/EthicsGuidelines}?
    \item[] Answer: \answerYes{}
    \item[] Justification: The study analyzes publicly released model checkpoints and established evaluation data, does not involve people, and reports limitations intended to prevent unsupported mechanistic or deployment claims.

\item {\bf Broader impacts}
    \item[] Question: Does the paper discuss both potential positive societal impacts and negative societal impacts of the work performed?
    \item[] Answer: \answerYes{}
    \item[] Justification: The Limitations and Broader Impact section discusses the value of more reliable monitoring and the risk that an overstated knowledge--reasoning distinction could encourage misleading competence claims.
    
\item {\bf Safeguards}
    \item[] Question: Does the paper describe safeguards that have been put in place for responsible release of data or models that have a high risk for misuse (e.g., pre-trained language models, image generators, or scraped datasets)?
    \item[] Answer: \answerNA{}
    \item[] Justification: The paper releases neither a new generative model nor a new scraped dataset and does not introduce an asset with a distinct high-risk misuse pathway.

\item {\bf Licenses for existing assets}
\item[] Question: Are the creators or original owners of assets (e.g., code, data, models), used in the paper, properly credited and are the license and terms of use explicitly mentioned and properly respected?
    \item[] Answer: \answerNo{}
    \item[] Justification: The appendix records the source-repository licenses and use terms for the evaluated models, datasets, and software. The public TriviaQA and MedQA mirrors do not state a license, so the study does not redistribute their content and the final release audit remains open.

\item {\bf New assets}
    \item[] Question: Are new assets introduced in the paper well documented and is the documentation provided alongside the assets?
    \item[] Answer: \answerNo{}
    \item[] Justification: The new extraction and analysis implementation is documented in the paper and repository, but it has not yet been packaged as an anonymous public submission asset with its final license.

\item {\bf Crowdsourcing and research with human subjects}
    \item[] Question: For crowdsourcing experiments and research with human subjects, does the paper include the full text of instructions given to participants and screenshots, if applicable, as well as details about compensation (if any)? 
    \item[] Answer: \answerNA{}
    \item[] Justification: The study involves no crowdsourcing and no research participants.

\item {\bf Institutional review board (IRB) approvals or equivalent for research with human subjects}
    \item[] Question: Does the paper describe potential risks incurred by study participants, whether such risks were disclosed to the subjects, and whether Institutional Review Board (IRB) approvals (or an equivalent approval/review based on the requirements of your country or institution) were obtained?
    \item[] Answer: \answerNA{}
    \item[] Justification: The study involves no crowdsourcing and no research participants.

\item {\bf Declaration of LLM usage}
    \item[] Question: Does the paper describe the usage of LLMs if it is an important, original, or non-standard component of the core methods in this research? Note that if the LLM is used only for writing, editing, or formatting purposes and does \emph{not} impact the core methodology, scientific rigor, or originality of the research, declaration is not required.
    \item[] Answer: \answerYes{}
    \item[] Justification: The Use of Large Language Models appendix states the role of general-purpose language-model tools and assigns responsibility for all scientific decisions and claims to the authors.

\end{enumerate}

\fi

\end{document}